\documentclass[letterpaper, 10 pt, conference]{ieeeconf}  

\IEEEoverridecommandlockouts                              

\usepackage{booktabs}
\usepackage{cite}
\usepackage{amsmath,amssymb,amsfonts}
\usepackage{algorithmic}
\usepackage{graphicx}
\usepackage{textcomp}
\usepackage{xcolor}
\usepackage{multirow}
\usepackage{booktabs}
\usepackage{microtype}

\title{\LARGE \bf
Making Parameterization and Constrains of Object Landmark Globally Consistent
via SPD(3) Manifold and Improved Cost Functions
}

\author{Yutong Hu$^1$ and Wei Wang$^{1,*}$
\thanks{This research is supported by the National Key Research and Development Program of China (2020YFB1313600) and Beijing Natural Science Foundation (3202015)}
\thanks{$^{1}$Authors are with the Robotics Institute, School of Mechanical Engineering and Automation, Beihang University, Beijing, China.}%
\thanks{$^*$Corresponding author, e-mail: wangweilab@buaa.edu.cn.}%
}

\begin{document}

\maketitle
\thispagestyle{empty}
\pagestyle{empty}

\begin{abstract}

Object-level SLAM introduces semantic meaningful and compact object landmarks that help both indoor robot applications and outdoor autonomous driving tasks. However, the back end of object-level SLAM suffers from singularity problems because existing methods parameterize object landmark separately by their scales and poses. Under that parameterization method, the same abstract object can be represented by rotating the object coordinate frame by 90$^{\circ}$ and swapping its length with width value, making the pose of the same object landmark not globally
consistent. To avoid the singularity problem, we first introduce the symmetric positive-definite (SPD) matrix manifold as an improved object-level landmark representation and further improve the cost functions in the back end to make them compatible with the representation. Our method demonstrates a faster convergence rate and more robustness in simulation experiments. Experiments on real datasets also reveal that using the same front-end data, our strategy improves the mapping accuracy by 22\% on average.

\end{abstract}

\section{Introduction and Related Work}
\setlength{\baselineskip}{11.5pt}
Compared with traditional SLAM based on sparse points, lines or planes, object-level SLAM provides a map that includes  higher-level and more compact landmarks embedded with semantic labels. Such human-like environment perception not only improves robustness under changing illuminations and viewpoints, but also enables the robot to conduct more complex actions, such as moving towards objects indoors and fetching them, or recognising cars and avoiding barriers outdoors. However, there are still singularity problems with the back end in object-level SLAM. We will review the previous work and then present the existing issues and our solutions.\\

\noindent\textbf{Oject-level SLAM:\ }The representation of semantic object landmarks in SLAM systems can be classified into several types, with a trade-off between fineness and computing complexity. The first attempt of object-level SLAM can be traced back to SLAM++\cite{salas-morenoSlamSimultaneousLocalisation2013a}. It used a CAD database to model chairs and tables in an office. In 2018, Martin et al. proposed
MaskFusion\cite{runzMaskFusionRealTimeRecognition2018}, representing each object by surfel cloud. With the help of deep learning based instance segmentation, MaskFusion no longer needs a prior CAD model database. However, this system requires heavy computational workload in order to support their dense model as well as the deep neural network. In 2019, CubeSLAM\cite{yangCubeSLAMMonocular3D2019} and QuadircSLAM\cite{nicholsonQuadricSLAMDualQuadrics2019} employed simpler and abstract geometries (Cubes or Ellipsoids) to represent object-level landmarks. Benefiting from their novel observation model, instead of the time-consuming segmentation on pixel level, only light-weight object detection boxes were needed. Although the modeled object may have a different shape compared with the real object, the essential information including position, orientation and occupied space is preserved, and both systems achieve real-time performance because of their simple and compact mathematical representation.

Recently, researchers have enhanced the mapping  accuracy of those object SLAM systems by proposing more robust landmark initialization method\cite{chenRobustDualQuadric2021,tianAccurateRobustObject2022} or adding more constraints to the landmarks, such as texture planes\cite{okRobustObjectbasedSLAM2019} and supporting planes\cite{hosseinzadehStructureAwareSLAM2019,yangMonocularObjectPlane2019}. Further attempts have also been made to extend the range of expressions of the object model. E.g., Zhen et al.\cite{zhenUnifiedRepresentationGeometric2021} unified planes, cylinders and other low-level landmarks as degenerate cases of quadrics, and Tschopp et al. \cite{tschoppSuperquadricObjectRepresentation2021} introduced super-ellipsoids to unify the representation of square and circular shaped objects.\\


\begin{figure}[t]
\centerline{\includegraphics[width=0.5\textwidth]{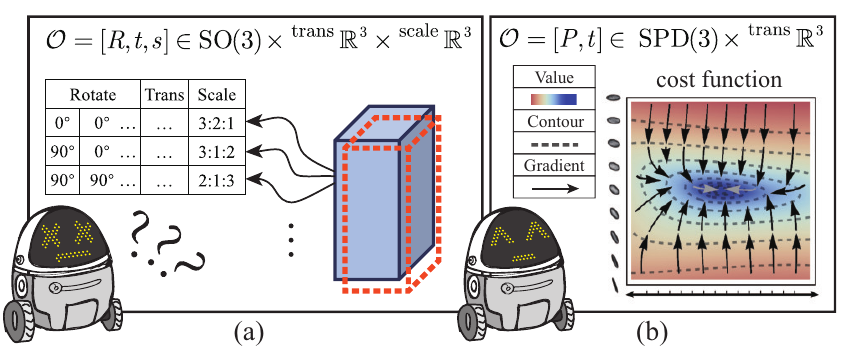}}
\caption{Demonstration of the singularity problem and different parameterization methods (a) Existing Rot-Trans-Scale method has more than one solution to the  same object (b) Our method utilize symmetric positive-definite matrix to avoid the singularity problem with better convergence.}
\label{head_fig}
\end{figure}

\noindent\textbf{Singularity in Object  Parameterization:\ }Although the geometric shapes of the cubes and ellipsoids are different, they both have 9 degrees of freedom (DOF) including \textbf{rotation}, \textbf{translation} and \textbf{scale} (of principal axis). Without loss of generality, we assume an object-level landmark $\mathcal{O}$ in the rest of this paper as an ellipsiod. The existing methods\cite{yangCubeSLAMMonocular3D2019,nicholsonQuadricSLAMDualQuadrics2019,zhenUnifiedRepresentationGeometric2021} consider an object landmark $\mathcal{O}$ to be in the direct product of the above three manifold spaces, i.e. $\mathcal{O} \in {\text{SO}}(3) \times {}^{{\text{trans}}}{\mathbb{R}^3} \times {}^{{\text{scale}}}{\mathbb{R}^3}$ or equivalently $\mathcal{O} \in {\text{SE}}(3) \times {}^{{\text{scale}}}{\mathbb{R}^3}$. However, such representation encounters the singularity problem: for the same object, there exists more than one solution to the sequence defined above. E.g., the object in Fig. \ref{head_fig} can be parameterizated as both $[0,0,0,0,0,0,3,2,1]$ and $[\pi / 2,0,0,0,0,0,3,1,2]$, which will influence the object SLAM back-end optimization process. The similar singularity problem 
appeared when representing a robot’s rotation in terms of Euler angles\cite{barfootStateEstimationRobotics2017}, and it was overcome by introducing the $\text{SO}(3)$ 
matrix Lie group (or unit quaternion, which is in the $\text{SU}(2)$ group). But for the case in object SLAM, the problem is not caused by rotation alone, for researchers have already introduced $\text{SO}(3)$ or $\text{SE}(3)$ in the aforementioned representation. Instead, it is \textbf{the coupling of rotation and scale} that makes ${\text{SO}}(3) \times {}^{{\text{trans}}}{\mathbb{R}^3} \times {}^{{\text{scale}}}{\mathbb{R}^3}$ not isomorphic to the manifold which the 9-DOF object landmarks actually lie on. \\

\begin{figure}[t]
\centerline{\includegraphics[width=0.5\textwidth]{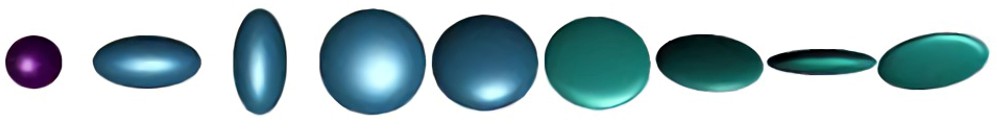}}
\caption{The symmetric positive definite matrix\cite{boumalManoptMatlabToolbox2014}. They can be visualized as ellipsoids: since the eigenvalues are all positive they can be taken as lengths of the axes while the directions are given by the eigenvectors.}
\label{spds_fig}
\end{figure}

\noindent\textbf{Optimization on Manifolds:\ } In the back end, the graph-based SLAM system solves an optimization problem on manifolds. To be precise, the optimization process is carried out by iterating in the \textbf{Tangent Space} and applying the \textbf{Retraction} back to the manifold. E.g., for a 3D robot pose $T$ in $\text{SE}(3)$, $\xi \in \text{se}(3)$ acts as the vector in tangent space and the matrix exponential mapping $\text{Exp}(\cdot)$ acts as the retraction. The iterating step can be written as:
\begin{equation}
    {T_{i + 1}} = \xi  \boxplus {T_i}\:{\text{ = }}\:{\text{Retract}}_{{T_i}}(\xi ) = \text{Exp} {\text{(}}{\xi ^ \wedge }{\text{)}}{T_i}
\end{equation}
where $\boxplus$ represents a generalization of the addition operation. The exponential mapping, although widely used, is not the only choice of retraction. E.g., as stated in \cite{dellaertFactorGraphsRobot2017},  sometimes it is computationally advantageous to forget about the Lie group structure and use other retractions defined in $\text{SE}(3)$ manifold. Furthermore, it is pointed out in \cite{chenCramerRaoBounds2021} that the choice of Riemannian metric of the manifold could affect the result accuracy and convergence rate in the process of back-end optimization, while it mainly focus on two choices of metric of $\text{SE}(3)$ in the case of solving 3D pose graphs. For object-level SLAM, the following parts of the paper will show the benefits of supplanting the classic Rot-Trans-Scale parameterization method by utilizing symmetric positive definite(SPD) matrix manifold. The $\text{SPD}(n)$ manifold, which can be visualized as Fig. \ref{spds_fig}, has been well studied over the past 20 years\cite{bhatiaPositiveDefiniteMatrices2007,moakherSymmetricPositiveDefiniteMatrices2006} and has been applied in covariance estimation\cite{drydenNonEuclideanStatisticsCovariance2009}, medical imaging and neuroscience\cite{pennecRiemannianFrameworkTensor2006}. However, to the best of our knowledge, they have not been applied in the field of SLAM.\\

 Our work introduces the SPD manifold into the back-end optimization process of object-level SLAM. And together with the improved cost function, we are able to avoid the above mentioned singularity problem during parameterization. The main contribution can be summarized as:
\begin{itemize}
\item We present how to optimize on the SPD manifold and how to integrate it into an optimization-based Object SLAM system
\item We improve several common-used constrained functions for object-level landmarks and compare two observation models of the ellipsoid.
\item Based on the above two points, an improved back end of monocular-camera Object SLAM system is proposed and experiments demonstrate that it outperforms the existing methods in the multi-constraint situation for estimating object landmarks.
\end{itemize}

The rest of this paper is structured as follows: Section \ref{one} discusses basic concepts about matrix, manifold and SPD(3), as well as how to integrate SPD(3) in optimization-based SLAM. Section \ref{two} covers the previously proposed measurement models and constraint models of object-level SLAM, and details our improved formulation of their cost functions. In Section \ref{three}, experiments in simulation and real datasets are presented. Finally, conclusions are drawn in Section \ref{four}.

\section{Basics about matrix, manifold,\\$\text{SPD}(3)$ and Object-level SLAM}\label{one}

\subsection{Matrix derivatives and inner products}

We start from the scalar-vector case $\frac{{\partial f}}{{\partial {\boldsymbol{x}}}}$, then vector-vector case $\frac{{\partial {\boldsymbol{f}}}}{\partial {\boldsymbol{x}}}$, and finally matrices case $\frac{{\partial {{\boldsymbol{F}}}}}{{\partial {{\boldsymbol{X}}}}}$: a vector $\boldsymbol{x}\in {\mathbb{R}^n}$ can be regarded as a matrix with only one column, i.e. $\boldsymbol{x}\in {\mathbb{R}^{n\times1}}$. If a scalar $f$ is a function of $\boldsymbol{x}$, then we have
\begin{equation}
 \mathrm{d}f = \sum\limits_{i = 1}^n {\frac{{\partial f}}{{\partial {x_i}}}\mathrm{d}{x_i} = } {\frac{{\partial f}}{{\partial {\boldsymbol{x}}}}^T}\mathrm{d}{\boldsymbol{x}}.   \label{df}
\end{equation}
Here we actually omit the matrix trace $\text{tr}(\cdot)$, and since we have $y=\text{tr}(y)$ for all $y \in {\mathbb{R}^{1 \times 1}}$, we can also write \eqref{df} as $\operatorname{tr}\left({\frac{\partial f}{\partial \boldsymbol{x}}}^{T} \mathrm{d} \boldsymbol{x}\right)$. For $f(\boldsymbol{X})$ where ${\boldsymbol{X}} \in {\mathbb{R}^{n \times m}}$, we have:
\begin{equation}
    \mathrm{d} f=\sum_{i=1}^{m} \sum_{j=1}^{n} \frac{\partial f}{\partial X_{i j}} \mathrm{d} X_{i j}=\operatorname{tr}\left({\frac{\partial f}{\partial \boldsymbol{X}}}^{T} \mathrm{d} \boldsymbol{X}\right)
\end{equation}
where $\frac{{\partial f}}{{\partial {\boldsymbol{X}}}} \in {\mathbb{R}^{m \times n}}$ and ${( {\frac{{\partial f}}{{\partial {\boldsymbol{X}}}}} )_{ij}} = \frac{{\partial f}}{{\partial {X_{ij}}}}$. For matrices $\boldsymbol{A}$, $\boldsymbol{B}$ with the same size, we have $\text{tr}(\boldsymbol{A}^T\boldsymbol{B})=\sum_{i,j}\boldsymbol{A}_{ij}\boldsymbol{B}_{ij}$, so $\text{tr}(\boldsymbol{A}^T\boldsymbol{B})$ here actually acts as an inner product $\langle {\boldsymbol{A}},{\boldsymbol{B}}\rangle $ of the two matrices and therefore unifies scalar-to-vector and scalar-to-matrix derivatives. As for vector function $\boldsymbol{f}(\boldsymbol{x})$, we simply line up the scalar-to-vector derivatives, i.e., ${\boldsymbol{J}} = \frac{{\partial {\boldsymbol{f}}}}{{\partial {\boldsymbol{x}}}} = {\left[ {\frac{{\partial {f_1}}}{{\partial {\boldsymbol{x}}}}, \cdots \frac{{\partial {f_n}}}{{\partial {\boldsymbol{x}}}}} \right]^T}$, and $\boldsymbol{J}$ is called the Jacobian matrix. However, in the case of matrix function $\boldsymbol{F}(\boldsymbol{X})$, $\frac{{\partial {{\boldsymbol{F}}^{m\times n}}}}{{\partial {{\boldsymbol{X}}^{p\times q}}}}$ cannot be written directly as a matrix. One common way to represent this derivative is Tenser, a generalized matrix. Another way is use $\text{vec}( \cdot) :{\mathbb{R}^{m \times n}} \to {\mathbb{R}^{mn}}$ to vectorize the two matrices, and then we can represent the derivative the  by a compatible Jacobi matrix:
\begin{equation}
    {\boldsymbol{J}^{mn \times pq}} = \frac{{\partial \:{{\left[ {{\operatorname{vec}(}{{\boldsymbol{F}}^{m \times n}}{)}} \right]}^{mn}}}}{{\partial \:{{\left[ {{\operatorname{vec}(}{{\boldsymbol{X}}^{p \times q}}{)}} \right]}^{pq}}}}.
\end{equation}
The above two forms can be transformed with one to another by operators such as the Kronecker product and
more detailed theory can be found in \cite{huMatrixCalculusDerivation2012} and \cite{petersenMatrixCookbook2008}. 

The point of the aforementioned concepts is to ensure that, \textbf{not having to be a vector, the matrix itself} can be stored and used as a state variable in the SLAM back-end optimization iteration process. To go further, because the real matrix group and many of its subsets are smooth manifold\cite{absilOptimizationAlgorithmsMatrix2009}, an inner product $\langle  \cdot , \cdot \rangle $ on their tangent space induces a Riemannian metric, making the manifold further a Riemannian manifold. E.g., an Euclidean space is a Riemannian manifold with the Riemannian metric $\langle A,B\rangle  = {\text{tr}}({A^T}B)$, which is essentially a choice of inner product for each element on the manifold.

\subsection{Using SPD(3) manifolds in object-level SLAM}\label{SPD_sec}

Modern SLAM back-end frameworks like g2o\cite{grisettiG2oGeneralFramework2011} or gtsam\cite{dellaertFactorGraphsGTSAM2012} allow user to customize the manifold on which the variables are lying, as well as its \textbf{Tangent spaces} and \textbf{Retractions}, and then automatically carries out the optimization process. In Quadric-SLAM, objects can be represented by a set of points on a quadratic surface
${{\boldsymbol{x}}_i} = {[x,y,z,1]^T} \in \{ {\boldsymbol{x}}|\ \Theta ({\boldsymbol{x}}) = 0\} $
, where:
\begin{equation}
\begin{aligned}
     \Theta ({\boldsymbol{x}}) = &A{x^2} + B{y^2} + C{z^2} +  2Dxy + 2Eyz +\\ &2Fxz + 2Gx + 2Hy + 2Iz + J   
\end{aligned}\label{ten}
\end{equation}

There are 10 parameters but only 9 degrees of freedom, for $\Theta ({\boldsymbol{x}}) = 0$ holds true if all terms are multiplied jointly by a factor. The shape function has a compact matrix form $\Theta ({\boldsymbol{x}}) = \boldsymbol{x}^T\boldsymbol{Q}\boldsymbol{x} = 0$, where:
\begin{equation}
\boldsymbol{Q}=\left[\begin{array}{cccc}
A & D & F & G \\
D & B & E & H \\
F & E & C & I \\
G & H & I & J
\end{array}\right]
\end{equation}
and $\boldsymbol{Q}$ need to be positive-definite to guarantee that the shape denoted by $\Theta$ is an ellipsoid. For this reason, a constrained dual-quadirc method was proposed \cite{nicholsonQuadricSLAMDualQuadrics2019}, while works such as \cite{zhenUnifiedRepresentationGeometric2021,okRobustObjectbasedSLAM2019} proposed a deeper understanding. Define the scaling matrix $\boldsymbol{S} = {\text{diag}}({s_1},{s_2},{s_3},1)$ , and the pose transformation matrix $\boldsymbol{T}(\boldsymbol{R},\boldsymbol{t})$. An ellipsoidal landmark $\boldsymbol{Q}$ in a world coordinate can be obtained by first scaling the unit ellipsoid ${{\boldsymbol{I}}^{4 \times 4}}$ at the origin point: ${\boldsymbol{S}^T\boldsymbol{I}}\boldsymbol{S}$ , and then transforming it by matrix $\boldsymbol{T}$, i.e. $\boldsymbol{Q}=\boldsymbol{T}^{-T} \boldsymbol{S}^{T} \boldsymbol{I S T}^{-1}$. Thus, $\boldsymbol{Q}$ can be decomposed as:
\begin{equation}
    \begin{aligned}
  {\boldsymbol{Q}} =  & {\left[ {\begin{array}{*{20}{c}}
  {{{\boldsymbol{R}}^T}}&{ - {{\boldsymbol{R}}^T}{\boldsymbol{t}}} \\ 
  {\boldsymbol{0}}&1 
\end{array}} \right]^T}\left[ {\begin{array}{*{20}{l}}
  {\boldsymbol{D}}&0 \\ 
  {\boldsymbol{0}}&1 
\end{array}} \right]\left[ {\begin{array}{*{20}{c}}
  {{{\boldsymbol{R}}^T}}&{ - {{\boldsymbol{R}}^T}{\boldsymbol{t}}} \\ 
  {\boldsymbol{0}}&1 
\end{array}} \right] \\ 
   =  & \left[ {\begin{array}{*{20}{c}}
  {{\boldsymbol{RD}}{{\boldsymbol{R}}^T}}&{ - {\boldsymbol{RD}}{{\boldsymbol{R}}^T}{\boldsymbol{t}}} \\ 
  *&{{{\boldsymbol{t}}^T}{\boldsymbol{RD}}{{\boldsymbol{R}}^T}{\boldsymbol{t}} + 1} 
\end{array}} \right] \\ 
\end{aligned} \label{decomp}
\end{equation}
where $\boldsymbol{D}=\text{diag}({s_1}^2,{s_2}^2,{s_3}^2)$, and the whole equation shows that an object can be parameterized by sequence $\mathcal{O} = [\boldsymbol{R},\boldsymbol{t},s_1,s_2,s_3] \in {\text{SE}}(3) \times {}^{{\text{scale}}}{\mathbb{R}^3}$ . However, as mentioned above, this way of parameterization has singularity problem. To solves this, we need to treat the block ${\boldsymbol{RD}}{{\boldsymbol{R}}^T}$ as its entirety, instead of decomposing it. Therefore, we introduce the $3\times 3$ symmetric positive definite matrix manifold\cite{boumalManoptMatlabToolbox2014} to give a globally
consistent representation of the same ellipsoid:
\begin{equation}
\text{SPD}(3) = \left\{\boldsymbol{P} \in \mathbb{R}^{3 \times 3}: \boldsymbol{P}=\boldsymbol{P}^{\top},\boldsymbol{P} \succ 0\right\}.
\end{equation}
$\boldsymbol{P} \succ 0$ here means $\boldsymbol{P}$ is positive definite, i.e. $a^{\mathrm{T}} \boldsymbol{P} a>0 \text { for all } a \in \mathbb{R}^{3} \backslash\{0\}$. As showed in Fig. \ref{spds_fig}, $\boldsymbol{P}$ can represent any ellipsoid with the center at the origin point of the coordinate\cite{linRiemannianGeometrySymmetric2019} , which further allows us to give a different way of decomposition of the dual-quadric:
\begin{equation}
    \begin{aligned}
  {\boldsymbol{Q}^*} =  & \left[ {\begin{array}{*{20}{c}}
  {{\boldsymbol{I}^{3 \times 3}}}&{ - \boldsymbol{t}} \\ 
  0&0 
\end{array}} \right]\left[ {\begin{array}{*{20}{c}}
  P&0 \\ 
  0&{ - 1} 
\end{array}} \right]{\left[ {\begin{array}{*{20}{c}}
  {{\boldsymbol{I}^{3 \times 3}}}&{ - \boldsymbol{t}} \\ 
  0&0 
\end{array}} \right]^T} \\ 
   =  & \left[ {\begin{array}{*{20}{c}}
  {\boldsymbol{P} - \boldsymbol{t}{\boldsymbol{t}^T}}&\boldsymbol{t} \\ 
  \boldsymbol{t}^T&{ - 1} 
\end{array}} \right] \\ 
\end{aligned} 
\end{equation}
i.e., to consider that the ellipsoid in the world coordinate is obtained by moving a particular ellipsoid $\boldsymbol{P}$ from the origin point through a translation $\boldsymbol{t}$. Then we have $\mathcal{O} = [\boldsymbol{P},\boldsymbol{t}] \in {\text{SPD}}(3) \times {}^{{\text{trans}}}{\mathbb{R}^3}$ without singularity. The \textbf{Tangent space} on SPD(3) manifold near $\boldsymbol{P}$ is $T_{p} \text{SPD}(3)=\left\{X \in \mathbb{R}^{3 \times 3} \mid X=X^{\mathrm{T}}\right\}$, which is quite natural because the differentiation of a symmetric matrix is also symmetric. However, in order to optimize on SPD(3), we cannot use the Euclidean metric $\text{tr}(\cdot,\cdot)$, under which the distance between positive definite and indefinite matrices is finite\cite{pennecRiemannianFrameworkTensor2006}, meaning that the iterative process may go 
out of the manifold. Instead, we use the linear affine metric \cite{moakherDifferentialGeometricApproach2005}: the metric of SPD matrices $\langle A,B\rangle $ near $P$ is ${\text{tr}}(P^{-1}{A}P^{-1}B)$ instead of ${\text{tr}}({A}B)$.

As for \textbf{Retraction}, we use the Exponential Retraction on SPD(3) manifold \cite{arsignyGeometricMeansNovel2007} :
\begin{equation}
    p \boxplus \xi  = {\text{Retrac}}{{\text{t}}_P}(\xi ) = {P^{\frac{1}{2}}}\operatorname{Exp} ({P^{ - \frac{1}{2}}}\xi {P^{ - \frac{1}{2}}}){P^{\frac{1}{2}}}
    \label{21}
\end{equation}
where $P^{\frac{1}{2}}$ is defined as follows: the positive definite symmetric matrix must be orthogonal-decomposable, i.e. $P = RER^T$, where:
\begin{equation}
E = \left[ {\begin{array}{*{20}{c}}
  {{s_1}^2}&0&0 \\ 
  0&{{s_2}^2}&0 \\ 
  0&0&{{s_3}^2} 
\end{array}} \right],{R^T} = {R^{ - 1}}. \label{svd}
\end{equation}Thus, in \eqref{21} we can let $P^{\frac{1}{2}} = R\sqrt{E}R^T$, where $\sqrt{E} = \text{diag}(s_1,s_2,s_3)$, so that it satisfies $P^{\frac{1}{2}}P^{\frac{1}{2}}=P$.  Notice that although there are more than one pair of $R,E$ satisfying $P = RER^T$, the results of the expressions $P^{\frac{1}{2}} = R\sqrt{E}R^T$ remain the same when substituted with each pair of $R,E$ . 

Finally, having all the properties and operations needed for SPD(3) been defined, we can use it in the back-end of object-level SLAM. The detailed optimization process on the matrix manifold using  G-N or L-M method can be found in \cite{absilOptimizationAlgorithmsMatrix2009} and  \cite{adlerNewtonMethodRiemannian2002}.

\section{Improvement of Cost Functions}\label{two}
To obtain a robust object landmark, measurement factors and a variety of semantic prior factors \cite{okRobustObjectbasedSLAM2019,liaoSOSLAMSemanticObject2021} are used in the back-end framework, expressed by error functions and cost functions. However, some of the existing formulation of those functions are not compatible with our proposed SPD(3) parameterization. Besides, as the number of constraint factors grows, these cost functions may conflict with each other \cite{liaoSOSLAMSemanticObject2021} accompanied by the singularity problem, leading to poor optimization results. To solve it, simply replacing the manifold space as in Sec. \ref{SPD_sec} is not enough, improving the formulation of these functions is also needed.
\begin{figure}[!t]
\centerline{\includegraphics[width=0.5\textwidth]{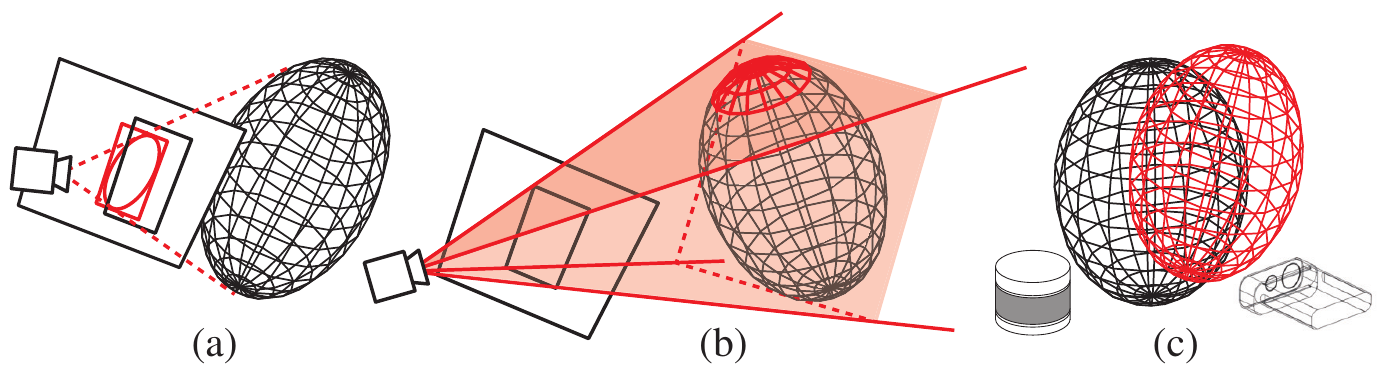}}
\caption{Object measurement models. (a) The inverse model for mono-camera (b) The semi-inverse model for mono-camera (c) Obtaining the entire ellipsoid in one measurement using RGB-D camera or 3D LiDAR}
\label{measure_fig}
\end{figure}
\subsection{Discussion of the measurement models}\label{sec:4a}
We first revise several state-of-the-art measurement models. As shown in Fig. \ref{measure_fig}, the process of getting an object detection box for a quadric-based SLAM using Deep Learning detectors can be interpreted in several ways, which we understand as inverse and semi-inverse measurement models.

The inverse model used in \cite{nicholsonQuadricSLAMDualQuadrics2019} and \cite{okRobustObjectbasedSLAM2019} assumes that the calculated bounding box $\bar B$ of an object landmark on the image should be the same as the result $B$ given by the object detector, as shown in Fig. \ref{measure_fig}(a). The bounding box of a dual-quadric $Q^*$ can be obtained by first getting the dual-conic $G^*$ of its projection onto the image:
\begin{equation}
    {{\boldsymbol{G}}^*} = {\boldsymbol{K}}\left[ {{{\boldsymbol{R}}_c}\mid {{\boldsymbol{t}}_c}} \right]{\boldsymbol{Q}}^*{\left[ {{{\boldsymbol{R}}_c}\mid {{\boldsymbol{t}}_c}} \right]^T}{{\boldsymbol{K}}^T}
\end{equation}
where $K$ is the camera intrinsic matrix and $R_c,t_c$ represent the camera pose. Then solve ${l_i}^T \boldsymbol{G}^* {l_i}=0$ for the four bounding box edges. Considering that the edges of the bounding box are either vertical or horizontal, i.e. $l_{u}=[1,0,-u]$ and $l_{v}=[0,1,-v]$, we can obtain the closed-form solution:
\begin{equation}
\begin{aligned}
&\hat{u}_{l}, \hat{u}_{r}=G_{1,3}^{*} \pm \sqrt{G_{1,3}^{*}{ }^{2}-G_{1,1}^{*} G_{3,3}^{*}}, \\
&\hat{v}_{u}, \hat{v}_{d}=G_{2,3}^{*} \pm \sqrt{G_{2,3}^{*}{ }^{2}-G_{2,2}^{*} G_{3,3}^{*}}.
\end{aligned}
\label{sqrt_measure}
\end{equation}
Thus, the measurement error function can be constructed as:
\begin{equation}
    f_{\text{box-inv}}(R_c,t_c,Q) = \bar B - B, \quad \bar B = [\hat{u}_{l },\hat{u}_{r},\hat{v}_{u},\hat{v}_{d}]. \label{inv}
\end{equation}

In contrast, the semi-inverse model used in \cite{liaoSOSLAMSemanticObject2021} assumes that the plane $\pi_i$ corresponding to the bounding box edge $l_i$ should be tangent to the object landmark, i.e. $\pi_i^T Q ^* \pi_i = 0$, as shown in Fig. \ref{measure_fig}(b), where $\pi_i = K[R_c|t_c]l_i$. Thus, another measurement error function can be constructed as:
\begin{equation}
    f_{\text{box-semi}}(R_c,t_c,Q) = \sum\limits_{i = 1}^4 {{\pi _i}^T{Q^*}} {\pi _i}. \label{semi}
\end{equation}

Object SLAM system\cite{nicholsonQuadricSLAMDualQuadrics2019} also used the semi-model to obtain an initial estimate using the SVD decomposition, but switched back to the inverse model in the rest of the optimization process. It is difficult to distinguish which model is better by theoretical analysis because \eqref{semi} has a simpler and more compact expression, while \eqref{inv} has a clearer meaning 
of Maximum a posteriori estimation. So we provide an experiment on how to choose them in Sec. \ref{sec:simu}.

In addition, although this paper focuses on the case of monocular cameras, it is worth mentioning that if use RGB-D camera or 3d LIDAR, the method proposed in \cite{zhenUnifiedRepresentationGeometric2021,liaoRGBDObjectSLAM2020} allows us to obtain the entire ellipsoid in each measurement and compare their errors directly, as in Fig. \ref{measure_fig}(c).

\begin{figure}[tb]
\centerline{\includegraphics[width=0.5\textwidth]{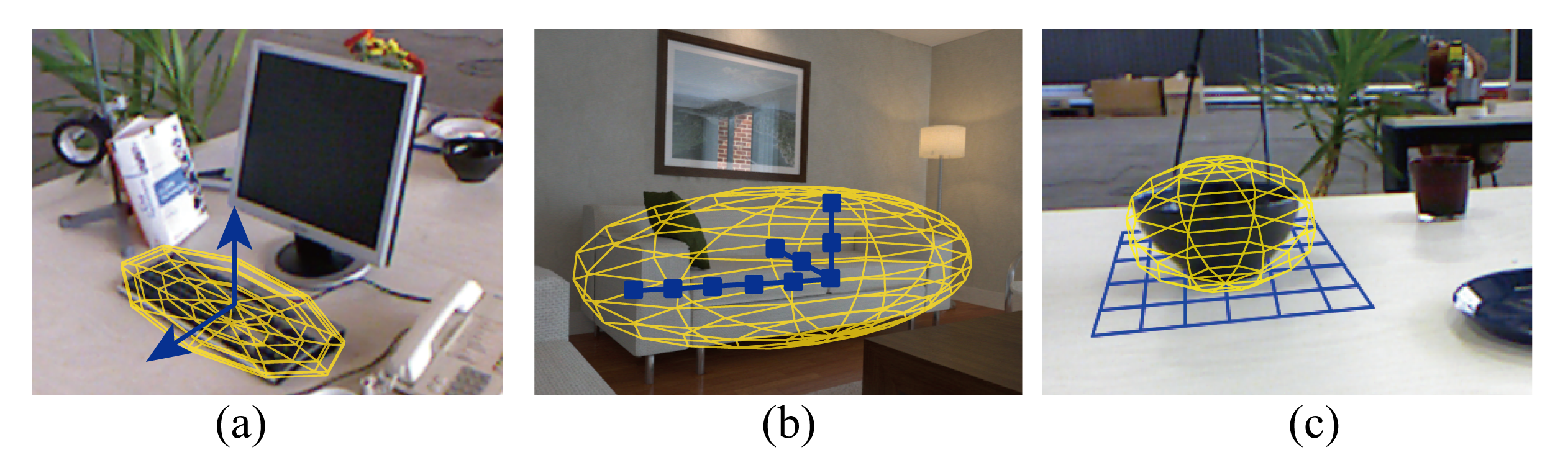}}
\caption{Constraints on objects. (a) Orientation constraint (b) Scale constraint (c) Supporting Plane constraint}
\label{cons_fig}
\end{figure}

\subsection{Improved representation of orientation constraints}
\begin{figure}[tb]
\centerline{\includegraphics[width=0.5\textwidth]{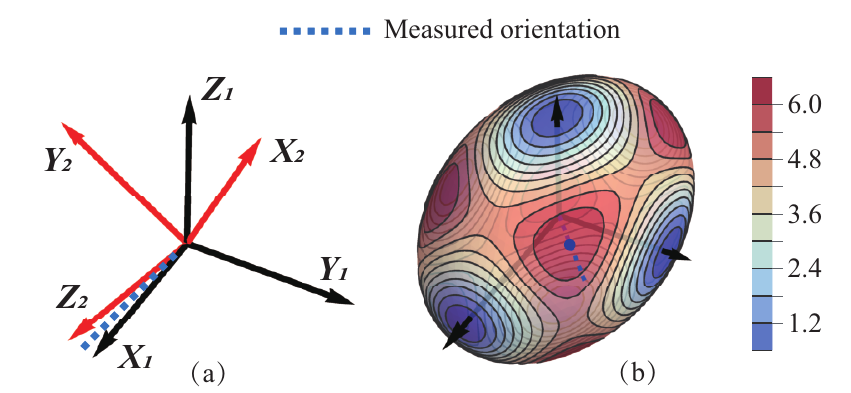}}
\caption{Details of orientation constraints. (a) The orientation of the object may face towards the same or opposite direction of one of the three axes (b) Visualization of the orientation cost function}
\label{ori_fig}
\end{figure}

Objects indoors, such as keyboards and chairs, are usually placed horizontally and have a clear orientation. Therefore, orientation factor has been further used to constrain object landmarks \cite{hosseinzadehStructureAwareSLAM2019,jablonskyOrientationFactorObjectoriented2018}, as shown in Fig. \ref{cons_fig}(a). However, for a given object coordinate frame, the orientation of the object may face towards the same or opposite direction of one of the three axes, as shown in Fig. \ref{ori_fig}(a), reflecting the singularity problem in another way.

We start our discussion with the relationship between the direction vector $\boldsymbol{n}$ of one of the three axes and the direction vector $\boldsymbol{m}$ of the supposed orientation. Ideally, $\boldsymbol{n}$ needs to be parallel \textbf{or} vertical to $\boldsymbol{m}$ which can be expressed as:
\begin{equation}
    (n \cdot m)\:{\text{or}}\:(n \times m)\:{\text{is}}\:0 \Leftrightarrow (n \cdot m)\:(n \times m) = 0.
\end{equation}
Inspired by this, the cost function can be constructed as:
\begin{equation}
    f_{\text{ori}}(Q) = \left[ {\begin{array}{*{20}{c}}
  {(R[:,1] \times m) \cdot (R[:,1] \cdot m)} \\ 
  {(R[:,2] \times m) \cdot (R[:,2] \cdot m)} \\ 
  {(R[:,3] \times m) \cdot (R[:,3] \cdot m)} 
\end{array}} \right]
\end{equation}
where $R$ can be obtained by SVD decomposition as in \eqref{svd} and $R[:,i]$ means the $i$-th column vector of $R$, representing the direction of the $X,Y,Z$ axes of the object frame. For a certain object landmark, the function $\lVert f_{\text{ori}} \rVert$ with respect to the variation of $m$ can be visualized by projecting the value onto a surface, as shown in Fig. \ref{ori_fig}(b). It can be seen that the function value is symmetric in all three directions $X,Y,Z$ and has six minimum points, corresponding to the same or opposite direction of the three axes. The function takes the maximum value when it is not close to any of the three axes. These properties ensure that for the same landmark, the value of the orientation constraint function is independent of the choice of the principal axes.

\subsection{Improved representation of scale constraints}
\setlength{\baselineskip}{11pt}
The scale constraint is another commonly used constraint \cite{yangCubeSLAMMonocular3D2019} \cite{okRobustObjectbasedSLAM2019}, which is particularly effective in the case of a straight-moving monocular camera because the depth unobservability problem\cite{chenRobustDualQuadric2021} can be overcome by knowing the shape of the object. The true scale can be set as a prior for each semantic label, and the measured scale $[s_1,s_2,s_3]$ also can be obtained by SVD decomposition as in \eqref{svd}. Notice that the SVD decomposition in the computer guarantees $s_1 \geq s_2 \geq s_3$, so the pre-defined prior needs to satisfy $a \geq b \geq c$ as well. Then we can split the scale cost function into two pieces, the shape term:
\begin{equation}
    {f_{\text{shape}}} = \left[ {\begin{array}{*{20}{c}}
  {{s_1}/{s_3}} \\ 
  {{s_2}/{s_3}} 
\end{array}} \right] - \left[ {\begin{array}{*{20}{c}}
  {a/c} \\ 
  {b/c} 
\end{array}} \right]
\end{equation}
and the size term:
\begin{equation}
    {f_{\text{size}}} = \det ({Q_{33}}) - a\cdot b\cdot c
\end{equation}
where $Q_{33}$ is the $3\times 3$ upper left submatrix of $Q$.

\subsection{Further Discussions}
As shown in Fig. \ref{cons_fig}(c), another commonly used constraint is the supporting plane constraint \cite{hosseinzadehStructureAwareSLAM2019,liaoRGBDObjectSLAM2020}. E.g., the cup is on the table and the sofa is on the floor. The mathematical formulation describing an object $Q$ on the plane $\pi$ is already good enough that there is no need to improve it:
\begin{equation}
    {f_{\sup }} = {\pi}^T{Q^*}{\pi}.
\end{equation}

As a consequence, assuming Gaussian measurement and process models, the entire object-level SLAM problem can be modeled as a nonlinear least-squares problem with pose constraints, measurement constraints, orientation constraints, supporting plane constraints and scale constraints:
\begin{equation}
\begin{gathered}
\{\hat{\mathcal{X}}\},\{\hat{\mathcal{O}}\}=\underset{\{\mathcal{X}\},\{\mathcal{O}\}}{\arg \min }\left(\sum_{i}\left\|f_{\text {pos }}(\mathcal{X})\right\|_{\Sigma_{p}}+\right. \\
\sum_{j}\left\|f_{\text {box }}(\mathcal{X}, \mathcal{O})\right\|_{\Sigma_{b}}+\sum_{k}\left\|f_{\text {ori }}(\mathcal{O})\right\|_{\Sigma_{o}}+\sum_{l}\left\|f_{\text {sup }}(\mathcal{O})\right\|_{\Sigma_{\pi}} \\
\left.+\sum_{m}\left\|f_{\text {shape }}(\mathcal{O})\right\|_{\Sigma_{s}}+\sum_{n} \left\|f_{\text {size }}(\mathcal{O}) \right\|_{\Sigma_{v}}\right)
\end{gathered}
\label{eq:fac-gra}
\end{equation}
where $\{ \mathcal{X}|\mathcal{X} \in {\text{SE}}(3)\} $ is the set of camera poses, $\{ \mathcal{O}|\mathcal{O} \in {\text{SPD}}(3) \times {}^{\text{scale}}{\mathbb{R}^3}\} $ is the set of object landmarks, and ${\lVert\cdot\rVert}_\Sigma$ is the Mahalanobis norm that directly scales
the error inversely proportional to the square
root of the covariance term $\Sigma$.

In the end, it is worth noting that those improved functions are actually compatible with the original ${\text{SO}}(3) \times {}^{{\text{trans}}}{\mathbb{R}^3} \times {}^{{\text{scale}}}{\mathbb{R}^3}$ manifold since they use matrix decomposition. The combined utilization of dot and cross products in the orientation constraint, and the sorting before the scale constraint can be implemented in object-level SLAM even if $\text{SPD}(3)$ manifolds are not used. 

\begin{figure}[tb]
\centerline{\includegraphics[width=0.45\textwidth]{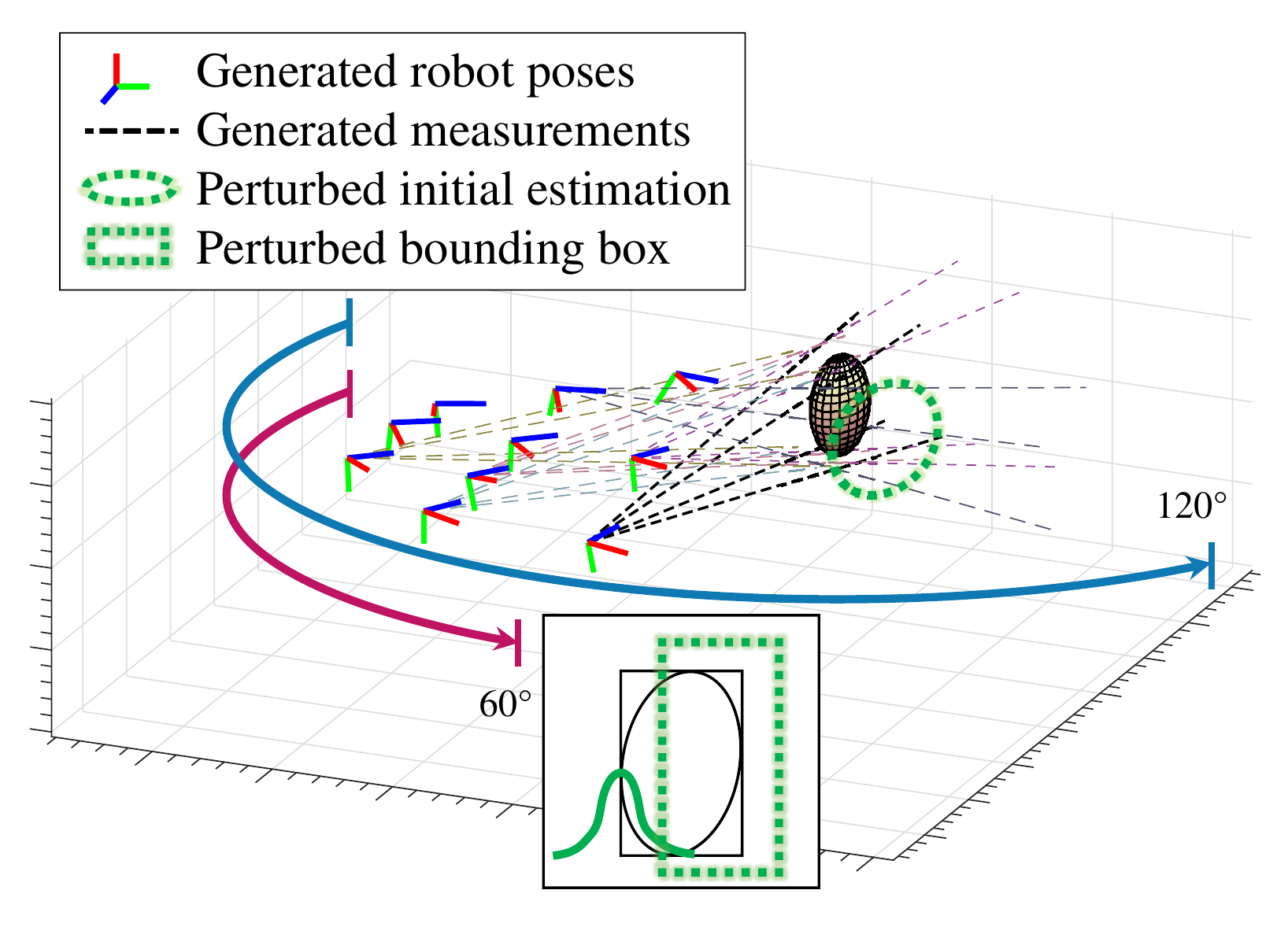}}
\caption{The synthetic environment. Camera frames are randomly generated within two angle ranges to simulate both insufficient and sufficient observation. The measured bounding box and initial estimate of landmark is perturbed by Gaussian noises.}
\label{simu_fig}
\end{figure}

\begin{figure*}[!t]
\centerline{\includegraphics[width=0.9\textwidth]{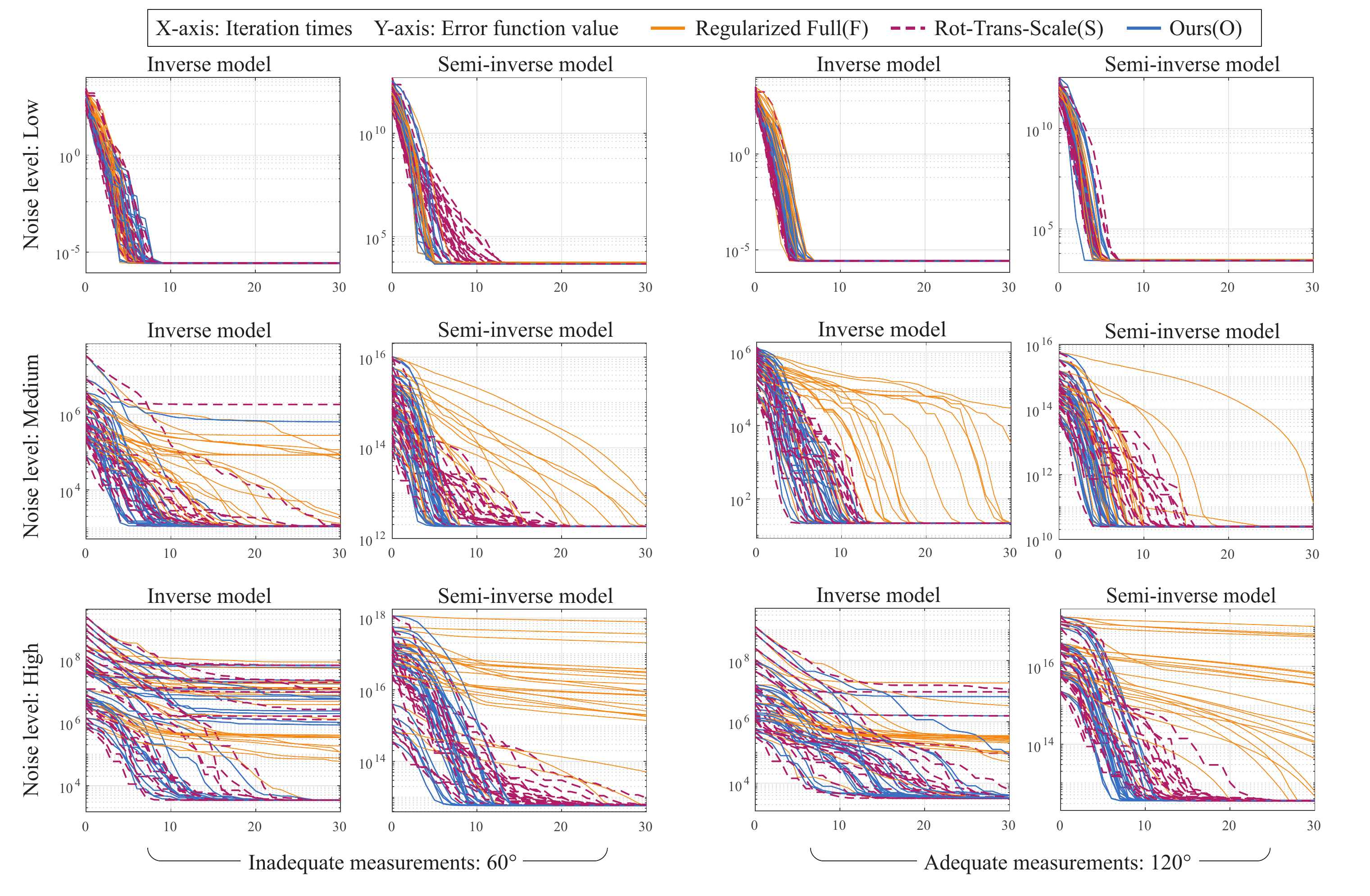}}
\caption{The convergence behavior of optimization process under various Noise level, Viewing angle range, Parameterization method and Measurement model. Our method outperforms others in convergence speed. The semi-inverse measurement model has higher robustness under large noise.}
\label{12_fig}
\end{figure*}

\section{Experiments}\label{three}
\begin{table}[tb]
\caption{Correspondence Between Covariance and Noise Levels}
\label{noise-tab}
\centering
\begin{tabular}{@{}lccc@{}}
\toprule
Noise Level     & Low                & Medium              & High             \\ \midrule
Box $\ \sigma_b$  & 0 pixel            & 5 pixel             & 10 pixel         \\
Init $\sigma_\theta,\sigma_t,\sigma_{\scriptscriptstyle{\%}}$ & $10^\circ$, 0.1m, 10\%  & $20^\circ$, 1m, 30\% & $40^\circ$, 3m, 50\%  \\ \bottomrule
\end{tabular}
\end{table}
Experiments are made in both synthetic environment and real datasets. The simulation  experiments focus on cases where only camera poses and measurement constraints are involved, to verify the effect brought by the introduction of SPD(3) and compare the two measurement models mentioned in Sec. \ref{sec:4a}. The experiments in real dataset further demonstrate the effect of combining SPD(3) with our improved cost functions.

\subsection{Simulation}\label{sec:simu}
\subsubsection{Data Generation}
We create an environment with only one ellipsoid landmark and  10 frames of its observation, as shown in Fig. \ref{simu_fig}. The ellipsoid is randomly generated in a bounded space ($1\text{m}\times2\text{m}\times3\text{m}$). Frames are randomly generated within two angle ranges: within $60^\circ$ to simulate a collection of insufficient observation, e.g., observed from a typical forward-moving robot \cite{okRobustObjectbasedSLAM2019}, and within $120^\circ$ to simulate a collection of sufficient observation. For each frame, we use \eqref{sqrt_measure} to generate its object detection box.
\subsubsection{Noise Generation}
Since the simulation experiments focus on object-level landmarks, we only consider noise that affects the landmark estimates. We add a Gaussian noise to the initial estimate of the object landmark according to the parameters $[\sigma_\theta,\sigma_t,\sigma_{\scriptscriptstyle{\%}}]$, where the pose is perturbed as ${T_i} = \exp ({\xi ^ \wedge }){\bar T_i},\xi  \sim \mathcal{N}\left( {{\mathbf{O}},\text{diag}({\sigma _\theta },{\sigma _\theta },{\sigma _\theta },{\sigma _t},{\sigma _t},{\sigma _t})} \right)$. And three main axes are perturbed as ${s_i}= \bar{s_i} + \mathcal{N}(0,\bar{s_i}\sigma_{\scriptscriptstyle{\%}}) $. We then add a Gaussian noise $\mathcal{N}(0,\sigma_b)$ to each edge of the observed object detection boxes as well. The exact value of three levels of noise: low(L), medium(M) and high(H) is defined in Table \ref{noise-tab}, which will be used to test the behaviors of different parameterizations.
\subsubsection{Implementation of back-end optimization}
We use three methods of lankmrak parameterization: the mentioned Rot-Trans-Scale (S) \cite{okRobustObjectbasedSLAM2019,zhenUnifiedRepresentationGeometric2021} method that represent an object by $\mathcal{O} = [\boldsymbol{R},\boldsymbol{t},\boldsymbol{s}] \in {\text{SO}}(3) \times {}^{{\text{trans}}}{\mathbb{R}^3} \times {}^{{\text{scale}}}{\mathbb{R}^3}$, our (O) method that represent an object by $\mathcal{O} = [\boldsymbol{P},\boldsymbol{t}] \in {\text{SPD}}(3) \times {}^{{\text{trans}}}{\mathbb{R}^3}$, as well as a Regularized Full (F) parameterization method simply representing the object as $\mathcal{O} = [A,B,C,D,E,F,G,H,I,J] \in {\mathbb{R}^{10}}$, and regularize it \cite{nicholsonQuadricSLAMDualQuadrics2019} after each iteration to ensure the object is still an ellipsoid. The last method serves as baseline. Another similar simulation is implemented \cite{zhenUnifiedRepresentationGeometric2021} and the results are in general concordance with ours. However, we simulate the case of a monocular camera, whereas their simulation contains more shapes observed by 3D-LiDAR.
\subsubsection{Results}
We generate the data 24 times under each combination of noise level and viewing angle range, and the optimization process is repeated with various parameterization method and measurement model. It turns out that in the testing case, the time consumptions under different configurations in each iteration step of the back-end optimization are similar ($2 \pm 0.1$ms on a single core of i5-9400@2.9GHz), so we demonstrate the variation of the cost function with respect to the number of iterations to report the convergence behavior of each optimization process, as in Fig. \ref{12_fig}. 

\textls[-25]{
We start with some obvious results: for the same noise and observation model, a wider range of viewing angles leads to faster convergence speed. At low noise levels, all parametrization methods work similarly well. The Full-Parameterization approach, on the other hand, begins to fail as the noise increases, while the other two manifold-based methods have a similar success rate. However, we do observe that our method has a faster convergence rate (23\% higher on average). Furthermore, the curves of our method also tend to have fewer variations on the way of descent, which may be explained by the fact that our parametrization of a landmark state is unique, rather than the Rot-Trans-Scale parametrization who has several way to represent the same landmark state. In addition, it is remarkable that the semi-inverse measurement model is more robust than the inverse measurement model. Especially in the case of large noise, even the two manifold-based methods start to fail under the inverse measurement model, but they can still iterate to the minimum error under the semi-inverse measurement model.
}

\begin{table}[t]
\centering
\caption{Count of success and Average IoU under various cases.}
\label{tab:sym-iou}
\begin{tabular}{cc|cccc}
\hline
\multicolumn{2}{c|}{} & inv-60        & semi-60           & inv-120       & semi-120          \\ \hline
\multirow{3}{*}{\begin{tabular}[c]{@{}c@{}}Success \\ Times:\\ F+S+O\end{tabular}} & \scriptsize{L} & 24+24+24    & 24+24+24    & 24+24+24    & 24+24+24    \\
                & \scriptsize{M}               & 15+23+23      & \textbf{20+24+24} & 20+24+24      & \textbf{21+24+24} \\
                & \scriptsize{H}                & 1+13+12       & \textbf{17+24+24} & 3+19+21       & \textbf{7+24+24}  \\ \hline
\multirow{3}{*}{\begin{tabular}[c]{@{}c@{}}Average\\ IoU:\\ F/S/O\end{tabular}}    & \scriptsize{L}  & .99/.99/.99 & .99/.99/.99 & .99/.99/.99 & .99/.99/.99 \\
                & \scriptsize{M}               & .76/.90/.88   & .79/.86/.88       & .87/.96/.95   & .73/.89/.89       \\
                & \scriptsize{H}                 & .15/.39/.37   & .43/.56/.57       & .21/.69/.76   & .30/.76/.76       \\ \hline
\multirow{3}{*}{\begin{tabular}[c]{@{}c@{}}Average\\ Success\\ IoU:\end{tabular}}  & \scriptsize{L} & 0.99        & 0.99        & 0.99        & 0.99        \\
                & \scriptsize{M}               & \textbf{0.91} & 0.84              & \textbf{0.95} & 0.89              \\
                & \scriptsize{H}                & \textbf{0.63} & 0.57              & \textbf{0.85} & 0.76              \\ \hline
\end{tabular}
\end{table}

\begin{table*}[t]
\centering
\caption{front-end object landmark measurement and back-end SLAM result on indoor datasets}
\label{tab:slam-iou}
\begin{tabular}{cc|ccc|ccc}
\hline
\multicolumn{2}{c|}{\textbf{Datasets}} &
  \multicolumn{3}{c|}{Front-end Input: IoU / Orient Error} &
  \multicolumn{3}{c}{Back-end Output: IoU / Orient Error} \\
Id &
  Objects &
  {\color[HTML]{000000} Cube-Edge} &
  {\color[HTML]{0E79B2} Quadric-SVD} &
  {\color[HTML]{BF1363} Quadric-Multi} &
  {\color[HTML]{0E79B2} Quadric-SLAM} &
  {\color[HTML]{BF1363} Multi-SLAM-RTS} &
  Multi-SLAM-SPD \\ \hline
ICL room2 &
  4 &
  {\color[HTML]{000000} 0.33 / \ \ -\ \ } &
  {\color[HTML]{0E79B2} 0.072 / 32.8} &
  {\color[HTML]{BF1363} \textbf{0.478 / 14.0}} &
  {\color[HTML]{0E79B2} 0.082  / 27.0 } &
  {\color[HTML]{BF1363} \underline{0.438} / 12.2 } &
  \textbf{0.506 / 11.8} \\
Fr1\_desk &
  13 &
  {\color[HTML]{000000} \ \ - \ \ / \ \ -\ \ } &
  {\color[HTML]{0E79B2} 0.066 / 45.0} &
  {\color[HTML]{BF1363} \textbf{0.106 / 12.6}} &
  {\color[HTML]{0E79B2} 0.071  / 37.0 } &
  {\color[HTML]{BF1363} 0.131  / \underline{13.2} } &
  \textbf{0.254 / 10.6} \\
Fr2\_desk &
  12 &
  {\color[HTML]{000000} \ \ - \ \ / \ \ -\ \ } &
  {\color[HTML]{0E79B2} 0.130 / 44.7} &
  {\color[HTML]{BF1363} \textbf{0.192 / \ 8.8}} &
  {\color[HTML]{0E79B2} 0.172  / 30.8 } &
  {\color[HTML]{BF1363} 0.334  / \ 8.7 } &
  \textbf{0.345 / \ 7.3} \\
Fr2\_dishes &
  4 &
  {\color[HTML]{000000} \ \ - \ \ / \ \ -\ \ } &
  {\color[HTML]{0E79B2} 0.118 / \ \ -\ \ } &
  {\color[HTML]{BF1363} \textbf{0.312} / \ \ -\ \ } &
  {\color[HTML]{0E79B2} 0.293  / \ \ -\ \ \ } &
  {\color[HTML]{BF1363} 0.375  / \ \ -\ \ \ } &
  0.375 / \ \ -\ \ \  \\
Fr3\_cabinet &
  1 &
  {\color[HTML]{000000} \textbf{0.46} / \ \ -\ \ } &
  {\color[HTML]{0E79B2} 0.254 / 20.0} &
  {\color[HTML]{BF1363} 0.344 / \ \textbf{1.4}} &
  {\color[HTML]{0E79B2} 0.255 / 18.3 } &
  {\color[HTML]{BF1363} \underline{0.317} / \ \textbf{0.9} } &
  \textbf{0.361} / \ 1.0 \\
Average &
  34 &
  {\color[HTML]{000000} \ \ - \ \ / \ \ -\ \ } &
  {\color[HTML]{0E79B2} 0.101 / 42.4} &
  {\color[HTML]{BF1363} \textbf{0.211 / 10.9}} &
  {\color[HTML]{0E79B2} 0.139  / 32.6 } &
  {\color[HTML]{BF1363} 0.272  / 10.9 } &
  \textbf{0.333 / \ 9.1} \\ \hline
\end{tabular}
\end{table*}

\begin{figure*}[th]
    \centering
    \includegraphics[width=0.9\textwidth]{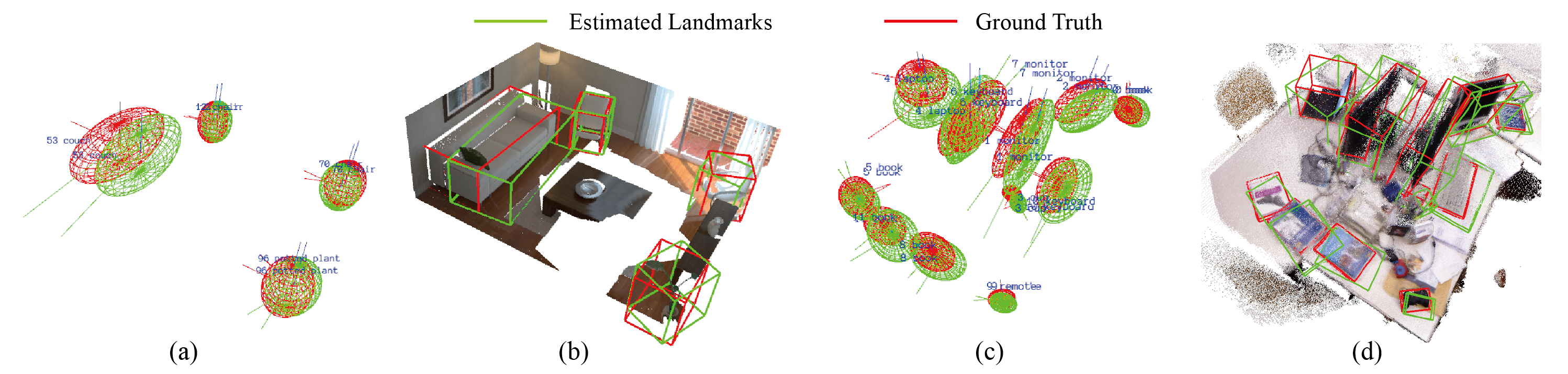}
    \caption{Semantic meaningful object-level maps provided by our improved back end. (a)(c) Object-level landmarks (b)(d) Landmarks with ground-truth point cloud, where objects are visualized as their circumscribed cubes to better demonstrate orientation.}
    \label{fig:b}
\end{figure*}

\setlength{\baselineskip}{11pt}
The back-end optimization is considered successful as long as the cost drops below a certain tolerance within the iterations. Yet, since the measurement error function is artificially designed, the convergence of the error does not necessarily imply that the optimization results of the object are close to the ground truth. Therefore, we calculate more quantitative data for each case, including total count of success and the Intersection-over-Union (IoU) between the result and ground truth, as in Table \ref{tab:sym-iou}. In addition, we find that the IoU of the three parameterization methods are similar as long as they succeed, so we jointly calculate the average IoU of the successful results as a whole to focus on the difference brought by the measurement model. We can see that although the inverse measurement model is less robust under high noise level, it can give a more accurate estimation of the object state when it succeeds. On the one hand, this inspires us to first get a landmark near the true value using the semi-inverse model and then switch to the inverse model for a more accurate result, as in \cite{nicholsonQuadricSLAMDualQuadrics2019}. On the other hand, it shows the importance of a good initial estimate in object-level SLAM. Each of \cite{okRobustObjectbasedSLAM2019,chenRobustDualQuadric2021,liaoSOSLAMSemanticObject2021,tianAccurateRobustObject2022} proposes their approach to get a better initial estimate.

\subsection{Experiments on Datasets}
We conduct experiments on both TUM RGB-D\cite{sturmEvaluatingEgomotionStructurefrommotion2012} and ICL-NUIM\cite{handaBenchmarkRGBDVisual2014} datasets, covering room-level and desktop-level environments.

\subsubsection{Implementation}
As in (21), the entire least-squares problem (or equivalently, the factor graph) in object-level SLAM contains camera pose constraints, measurement constraints, orientation constraints, size constraints, scale constraints, and plane constraints. We obtain the camera pose constraints by ORB-SLAM2 \cite{mur-artalOrbslam2OpensourceSlam2017} without loop closures, where depth information is used to provide a consistent scale for odometry and does not yet support the estimation of landmark parameters. As for those remaining constraints, our previous work \cite{liaoSOSLAMSemanticObject2021} proposed a method to obtain them at the front end using only RGB frames. In brief, we perform object detection with YOLOv3 to get bounding boxes, which we then formulate as semi-inverse measurement errors \eqref{inv}. And we further provide support planes, scales, frontside and topside orientations as object constraints.

\subsubsection{Details}

Since we are mainly working on back-end improvements, we take the same constraints obtained by our mentioned front end, and show the results given by back end without and with our improvements. To compare with the state-of-the-art algorithms, we cited the result from CubeSLAM\cite{yangCubeSLAMMonocular3D2019} and reproduced the performance of QuadricSLAM. We use the indicators Averange IoU and Averange Orientation Error to fully evaluate the mapping effects. The IoU evaluates the Intersection over Union between their circumscribed cubes of estimated object and ground-truth object. For objects with clear directions (e.g., Globe and Bowl are not included), Orientation Error (deg) evaluates the minimum rotation angle required to align the estimated object's three rotation axes with any axis of the ground-truth object to a straight line. In addition, there are more than one object in each scene, so we need to associate the detection boxes with landmarks. There has been works \cite{bowmanProbabilisticDataAssociation2017,dohertyProbabilisticDataAssociation2020} as well as our previous work \cite{liaoRGBDObjectSLAM2020} concentrating on this problem. As data association is not the focus of this paper, we use manually annotated data association to testify the best effectiveness in the experiments, which is the same as \cite{nicholsonQuadricSLAMDualQuadrics2019}.
\subsubsection{Results} Comparisons of the IoU and orientation errors before and after back-end optimization using various methods are given in Table \ref{tab:slam-iou}. Multi-SLAM-RTS introduces more constraints to Quadric-SLAM  while keeping the back-end representation of the quadric landmarks unchanged. Multi-SLAM-SPD further switches on our proposed object landmark parameterization and constraints formulation method. In the front-end initailization, due to the incorporation of multimodal constraints, the Quadric-Multi obtained significantly improved IoU and orientation of landmarks compared to the Quadric-SVD method. And after the back-end optimization, the results with the multimodal constraints are, as expected, still better than the original Quadric-SLAM. However, in contrast to Quadric-SLAM's back-end results, which always become more accurate than their SVD-initial results, our previous results (Multi-SLAM-RTS) sometimes become rather worse after back-end optimization. These unexpected results implies the existence of a conflict between the inappropriate constraint formulation and landmark parameterization.

\setlength{\baselineskip}{11pt}
Using the same front-end data, we replace the object parameters and cost functions of the back-end optimization with those proposed in this paper. The results (Multi-SLAM-SPD) show that our methods outperform the previous methods and, as conflicts no longer existed, have a 22\% increase in the average IoU and a 16\% decrease in the orientation error from the back-end results. 

As for trajectory accuracy, we do not find a significant  improvement when adding object landmarks into ORB-SLAM2, which is similar to the conclusion of 
\cite{nicholsonQuadricSLAMDualQuadrics2019,tianAccurateRobustObject2022,liaoSOSLAMSemanticObject2021}. We attribute this to the fact that the ORB-points based localization is already accurate enough in the indoor static environment, while the number of object landmarks is relatively small compared with the points, and the errors of detection are sometimes non-Gaussian. Therefore, back-end improvements can hardly solve the above problem. Although our refinements on object-level landmarks mainly bring benefits of perception in this paper, they are expected to bring long-term localization robustness during dynamic-environment SLAM in the future.



\section{Conclusion and Future Work}\label{four}
\setlength{\baselineskip}{11pt}
In summary, our paper introduces the SPD (3) manifold along with improved cost functions in Object-level SLAM to solve the singularity problem in both object landmark parameterization and constraints formulation. 

Experiments demonstrate that under mono RGB-camera observation, our method has comparable success rate as the classic Rot-Trans-Scale parameterization method, but converges faster when only object bounding boxes are provided. And when multiple constraints on objects are provided, such as orientation and scale constraints, it has higher accuracy of object landmark mapping.

To the best of our knowledge, SPD(3) is first introduced into the field of object-level SLAM, and we have only made a simple application of its Exponential Retraction, acting as addition operation $\boxplus$ during back-end optimization. More of its properties can be explored to improve the performance of object-level SLAM. E.g., denote the matrix logarithm as $\text{Log}(\cdot)$, the Logarithmic Map\cite{moakherDifferentialGeometricApproach2005} of $p,q\in\text{SPD}(3)$ is:
\begin{equation*}
    \log_p q = p \boxminus q = 
p^{\frac{1}{2}}\operatorname{Log}(p^{-\frac{1}{2}}qp^{-\frac{1}{2}})p^{\frac{1}{2}}\tag{22}
\end{equation*}
which naturally describes the difference between two ellipsoidal objects, and thus has great potential for object-level data association and relocalization.




\linespread{0.85}
\bibliography{SPD-IROS-refer.bib} 
\bibliographystyle{IEEEtran}  

\addtolength{\textheight}{-12cm}   

\end{document}